\documentstyle[aaai,a4,graphics]{article}

\title{A Symbolic and Surgical Acquisition of Terms\\through
Variation\thanks{\hspace*{1mm}We would like to thank the French
scientific documentation center {\em INIST/CNRS} for providing us with
data. All the experiments reported in this paper have been performed
on [Pascal] a list of 71,623 multi-domain terms and [Medic] a
1.56-million word medical corpus composed of abstracts of scientific
papers owned by {\em INIST/CNRS}. Many thanks to Jean Royaut\'e of
{\em INIST/CNRS} for his helpful and friendly collaboration. This work
has also benefited from rich discussions in the research group {\em
Terminologie et Intelligence Artificielle} of the {\em PRC
Intelligence Artificielle}.}}

\author{Christian Jacquemin\\Institut de Recherches en Informatique de
Nantes (IRIN)\\IUT de Nantes, 3 rue M$^{\mathrm{al}}$ Joffre, 44041
Nantes, France\\Phone~: +33 40 30 60 52\ ---\ Fax~:  +33 40 30 60
53\\\texttt{jacquemin@irin.iut-nantes.univ-nantes.fr}}

\begin{document}

\maketitle

\vspace*{-6.5cm}
\parbox{\textwidth}{In Proceedings, {\em Workshop``New Approaches to
Learning for NLP'' at 14th International Joint Conference on
Artificial Intelligence (IJCAI'95)}, Montr\'eal. {\em Forthcoming}.}
\vspace*{+5.2cm}

\begin{abstract}
Terminological acquisition is an important issue in learning for NLP
due to the constant terminological renewal through technological
changes. Terms play a key role in several NLP-activities such as
machine translation, automatic indexing or text understanding. In
opposition to classical once-and-for-all approaches, we propose an
incremental process for terminological enrichment which operates on
existing reference lists and large corpora. Candidate terms are
acquired by extracting variants of reference terms through {\em
FASTR}, a unification-based partial parser. As acquisition is
performed within specific morpho-syntactic contexts (coordinations,
insertions or permutations of compounds), rich con\-cep\-tual links
are learned together with candidate terms. A clustering of terms
related through coordination yields classes of conceptually close
terms while graphs resulting from insertions denote generic/specific
relations. A graceful degradation of the volume of acquisition on
partial initial lists confirms the robustness of the method to
incomplete data.
\end{abstract}

\section{Aims}

Multi-word terms and compounds play an increasing role in language
analysis for the following reasons~: their interpretation is rarely
transparent, they generally denote a specific class of mental or
real-world objects and the words composing them are strongly related.
Therefore, a correct processing of terms ensures a higher quality in
several applications of Natural Language Processing (NLP). In Machine
Translation, their lack of transparency makes word-for-word
translation fail and calls for specific descriptions. In Information
Retrieval, their high informational content makes them good
descriptors \cite{LewCroft90}. In parsing, the selectional
restrictions found between head words and their arguments within a
term give important clues for structural noun phrase disambiguation
\cite{Resnik93}.

As terms mirror the concepts of the domain to which they belong, a
constant knowledge evolution leads to a constant term renewal. Thus
terminological acquisition is a necessary companion to NLP,
specifically when dealing with technical texts. 

Tools for terminological acquisition, whether statistical, such as
\cite{ChurchHanks89}, or symbolic, such as \cite{Bourig93}, acquire
terms from large corpora through a once-and-for-all process without
consideration for any prior terminological knowledge. This lack of
incrementality in acquisition has the following drawbacks~:
\begin{itemize}
\item
Acquired terms must be merged with the initial ones with consideration
of eventual variants.
\item
Acquired terms are neither conceptually nor linguistically related to
the original ones.
\item
The set of original terms is ignored although it could be a useful
source of knowledge for acquisition.
\end{itemize}

It is possible to conceive a finer approach to term acquisition by
considering the local variants of terms within corpora. As term
variants generally involve more than one term, their extraction can
fruitfully exploit existing lists of terms in a process of non massive
incremental acquisition. For example, if {\em viral hepatitis} is a
known term, {\em viral and autoimmune hepatitis} is a variant of this
term (a coordination) which displays {\em autoimmune hepatitis} as a
candidate term. Moreover, this coordination indicates a strong
closeness between the interpretation of both terms which can be
associated to a link within a thesaurus. Henceforth, potential terms
acquired through acquisition techniques will be called {\em candidate
terms}. The decision whether to include a candidate term into a
terminology is outside the scope of our work.

\section{Acquiring with a Concern for Prior Knowledge}

Tools for acquiring terms generally operate on large corpora using
various techniques to detect term occurrences. There are mainly two
families of tools for term acquisition~: statistical measures and NLP
symbolic techniques.

The first family which comprises most of the tools is composed of
statistical analyzers which have little or no linguistic knowledge.
These applications take advantage of the specific statistical behavior
of words composing terms~: words which are lexically related tend to
be found simultaneously more frequently than they would be just by
chance. Pure statistical methods such as \cite{ChurchHanks89} are
rare. Generally some linguistic knowledge is associated to the
statistical measures through a prior \cite{Dail94} or a posterior
\cite{Smad93} filtering of correct syntactic patterns. The assumption
implicitly stated by statistical works, and which is backed up by our
study, is that it is more likely to find a term in the neighborhood of
another one than anywhere else in a text. More specifically, we assume
that the best way of combining two terms syntactically and
semantically is to build a specific structure that we call a {\em
variant} which is either a term or a restricted noun phrase and which
is observed within a small span of words.

The second approach to term acquisition consists of knowledge-based
methods which rely on local grammars of noun phrases and compounds
\cite{Bourig93}. Word sequences accepted by these grammars are
extracted through a more or less shallow parse of corpora and are good
candidate terms.

The counterpart of both sta\-tis\-ti\-cal and know\-le\-dge-ba\-sed
acqui\-si\-tions is to provide the user with large lists of candidates
which have to be manually filtered out. For example, {\em LEXTER}
\cite{Bourig93} extracts 20,000 occurrences from a 200,000-word corpus
which represent 10,000 candidate terms. It is due to a lack of initial
terminological knowledge along with a lack of consideration for
terminological variation that such methods propose too large sets of
terms. In order to reduce the volume of acquisition and also to
propose candidates which are more likely to be terms, this paper
presents a method based on an initial list of terms called {\em
reference terms}. The acquisition procedure starts from this supposed
comprehensive set of reference terms. It decomposes variations of
these terms found in corpora 
and is then able to detect candidate terms.

\subsection*{Updating Rather Than Acquiring}

Is it realistic to suppose that lists of terms exist for technical
domains~? The ever-growing mass of electronic documents calls for
tools for accessing these data which have to make extensive use of
term lists as sources of indexes. For this purpose, and for other
activities related to textual databases, more and more thesauri exist.
Some of them, such as the {\em Unified Medical Language System}
meta-thesaurus, carry conceptual and/or linguistic information about
the terms they contain. In our experiment we have used the [Pascal]
terminological list composed of 71,623 multi-domain terms without
conceptual links, provided by the documentation center {\em
INIST/CNRS}.

Because of the availability of large term lists, it is natural to lay
a greater stress on the updating of such data than on their
acquisition from scratch. Therefore, our approach to acquisition
focuses on how to improve a list of terms through the observation of a
corpus. Our approach also differs from previous experiments on term
acquisition because it yields conceptual links between candidate and
reference terms. It can be used to check or to enhance the conceptual
knowledge of thesauri in a way complementary to automatic semantic
clustering of terms through an observation of their syntactic contexts
\cite{Gref94}.

\section{A Micro-syntax for an Accurate Extraction}

The first step in our approach to terminological acquisition is the
extraction of term variants from a large corpus. The tool used is {\em
FASTR}, a unification-based partial parser. {\em FASTR} recycles lists
of reference terms by transforming them into grammar rules. Then, it 
dynamically builds term variant rules from these term rules. The
parser is described in \cite{Jacq94b} and, here, it will
just be sketched out, by focusing on the features that are relevant
for terminological acquisition. More specifically, we will omit the
aspects of the parser concerning its optimization and the feature
structures associated with rules and meta-rules.

\stepcounter{equation}\newcounter{formserAlb}\setcounter{formserAlb}{\theequation}
In such a simplified framework, each reference term corresponds to a
{\em PATR-II}-like rule \cite{Shieb86} comprising a context free
skeleton and lexical items. For example, rule~(\theformserAlb) denotes
the term {\em serum albumin} with a {\em
$\langle$Noun$\rangle$\,$\langle$Noun$\rangle$} structure~:

\vskip 3mm
{
{\begin{tabular}[t]{ll}
Rule&$ N_1 \rightarrow N_2\ N_3\ :$\\
&$<N_2\ lemma>\ \doteq\ ${\em `serum'}\\
&$<N_3\ lemma>\ \doteq\ ${\em `albumin'}.
\end{tabular}}
\hfill(\theformserAlb)
}
\vskip 3mm

\stepcounter{equation}\newcounter{formmeat}\setcounter{formmeat}{\theequation}
\stepcounter{equation}\newcounter{formserAlbTransf}\setcounter{formserAlbTransf}{\theequation}
At a higher level, a set of meta-rules operates on the term rules and
produces new rules describing potential variations. Each meta-rule is
dedicated to a specific term structure and to a specific type of
variation. For the sake of clarity, meta-rules are divided into two
sets -- meta-rules for two-word terms and meta-rules for three-word
terms -- and each set is subdivided into three subsets -- meta-rules
for coordination, insertion and permutation. Meta-rules for terms of
four words or more are ignored because they produce very few variants
(approximately 1\,\% of the variants). Meta-rule~(\theformmeat)
applies to rule~(\theformserAlb) and yields a new
rule~(\theformserAlbTransf)~:

\vskip 3mm
{
{\begin{tabular}[t]{ll}
Metarule&$ Coor(X_1 \rightarrow X_2\ X_3)$\\
&$\equiv\ \ X_1 \rightarrow X_2\ C_4\ X_5\ X_3\ :.$
\end{tabular}}
\hfill(\theformmeat)
}
\vskip 3mm
{
{\begin{tabular}[t]{ll}
Rule&$ N_1 \rightarrow N_2\ C_4\ X_5\ N_3\ :$\\
&$<N_2\ lemma>\ \doteq\ ${\em `serum'}\\
&$<N_3\ lemma>\ \doteq\ ${\em `albumin'}.
\end{tabular}}
\hfill(\theformserAlbTransf)
}
\vskip 3mm

\noindent This transformed rule accepts any sequence {\em serum $C_4\
X_5$\ albumin} as a variant of {\em serum albumin} where $C_4$\ is any
coordinating conjunction and $X_5$\ any single word. For example, it
correctly recognizes {\em serum and egg albumin} as a variant of {\em
serum albumin}. The second column of Table~\ref{tabacqui} presents
some other meta-rules for two-word terms together with examples of
pairs composed of a term and one of its variants. Currently, the
meta-grammar of {\em FASTR} for English includes 73 meta-rules for 2-
and 3-word terms~: 25 coordination meta-rules, 17 insertion
meta-rules and 31 permutation meta-rules (plus 66 meta-rules for
4-word terms which are not used for acquisition).

When term variants are described through meta-rules as in {\em FASTR},
it is very simple to devise a process for term acquisition~: each
paradigmatic meta-rule (or skeleton of a filtering meta-rule) is
linked to a pattern extractor, yielding a candidate term. As no
further analysis of the variants is required, such an acquisition is
extremely fast. The acquisition of terms by extracting patterns from
variants is processed as follows for the different categories of
variants~:
\begin{itemize}
\item
{\em Coordination.} The candidate term is the term coordinated with
the original one.
\item
{\em Insertion.} The candidate term is the term which has replaced the
head of the original term through substitution.
\item
{\em Permutation.} In a permutation of a 2-word term, the argument of
the original term is shifted from the left of the head to its right
and is transformed into a prepositional phrase. The candidate term is
the noun phrase inside this prepositional phrase. This definition is
extended to terms of 3 words or more where one of the arguments is
permuted.
\end{itemize}
The third column of Table~\ref{tabacqui} exemplifies patterns of
acquisition for each of the three categories of term
variants.

\begin{table*}[t]
\centering
\mbox{
\begin{tabular}[t]{|l|l|l|}
                            \hline
&Meta-rule and associated variant&Acquisition\\
                            \hline
                            \hline
Coordination&$X_2\ X_3\mapsto X_2\ X_4\ C_5\ X_3$&$X_2\ X_4$\\
(25 meta-rules)&{\em surgical closure}&{\em surgical exploration}\\
&\hskip 0.5cm $\mapsto$\ {\em surgical exploration and closure}&\\
                            \hline
                            \hline
Insertion&$X_2\ X_3\mapsto X_2\ X_4\ X_3$&$X_4\ X_3$\\
(17 meta-rules)&{\em medullary carcinoma}&{\em thyroid carcinoma}\\
&\hskip 0.5cm $\mapsto$\ {\em medullary thyroid carcinoma}&\\
                            \hline
                            \hline
Permutation&$X_2\ X_3\mapsto X_3\ P_4\ X_5\ X_2$&$X_5\ X_2$\\
(31 meta-rules)&{\em control center}&{\em disease control}\\
&\hskip 0.5cm $\mapsto$\ {\em center for disease control}&\\
                            \hline
\end{tabular}
}
\caption{\label{tabacqui}Acquisition through pattern extraction from
variants. (Examples are from [Medic].)}
\end{table*}

This method for term acquisition does not systematically succeed for
each encountered term variant. Some correct variants involve only one
term instead of two or more and cannot produce new candidates. For
example, {\em cells and their subpopulations} is a coordination
variant of {\em cell subpopulation} which is unproductive compared
with the variant exemplified for coordination in
Table~\ref{tabacqui}. Moreover, terms acquired through a variation may
already be reference terms (see the non-underlined candidates in
Tables~\ref{tabCoor}, \ref{tabIns} and \ref{tabPerm}). For the
reference list to be sufficiently comprehensive, it is expected and
even desirable that some of the acquired terms are already known.
Moreover, ``acquisitions'' of known terms are not useless because they
reveal conceptual links between these terms.

\section{Acquiring Conceptual Classes}

Tables~\ref{tabCoor}, \ref{tabIns} and \ref{tabPerm} exemplify some
terms acquired through the three main kinds of variations observed for
English~: coordinations, insertions and permutations. The terms
acquired through permutations are not conceptually related to the
original ones due to the syntagmatic nature of this transformation.
On the contrary, coordination and insertion variations relate semantically
close terms. We examine in turn the decomposition of these two kinds
of variations in the aim of acquiring conceptual links.

\begin{table}[t]
\centering
\mbox{
\begin{tabular}[t]{|ll|}
\hline
Candidate term&Reference Term\\
\hline
\hline
{\em abdominal aorta}&{\em Thoracic aorta}\\
$\underline{\textrm{\em acidic lipid}}$&{\em Neutral lipid}\\
$\underline{\textrm{\em active phase}}$&{\em Latent phase}\\
{\em adrenal gland}&{\em Thyroid gland}\\
{\em affective disorder}&{\em Cognitive disorder}\\
{\em aged animal}&{\em Young animal}\\
$\underline{\textrm{\em agonist bromocriptine}}$&{\em Agonist
antagonist}\\
{\em air conduction}&{\em Bone conduction}\\
$\underline{\textrm{\em amniotic fluid estimation}}$&{\em Ratio
estimation}\\
{\em aortic arch}&{\em Aortic coarctation}\\
{\em aortic valve}&{\em Mitral valve}\\
$\underline{\textrm{\em arterial acid base}}$&{\em Arterial blood}\\
\hline
\end{tabular}
}
\caption{\label{tabCoor}Examples of term acquisition through
coordination from [Medic]. Terms which do not belong to the reference
list are underlined.}
\end{table}

\begin{table}[t]
\centering
\mbox{
\begin{tabular}[t]{|ll|}
\hline
Candidate term&Reference Term\\
\hline
\hline
$\underline{\textrm{\em abdominal spear injury}}$&{\em Penetrating
injury}\\
$\underline{\textrm{\em ablating tool}}$&{\em Cutting tool}\\
{\em absorbed dose}&{\em Radiation dose}\\
$\underline{\textrm{\em access pressure}}$&{\em Blood pressure}\\
{\em accessory nerve}&{\em Spinal nerve}\\
$\underline{\textrm{\em acetylcholine receptor}}$&{\em Muscarinic
receptor}\\
$\underline{\textrm{\em acetylcholine receptor}}$&{\em Nicotinic
receptor}\\
$\underline{\textrm{\em acid analysis}}$&{\em Organic analysis}\\
$\underline{\textrm{\em acid base disorder}}$&{\em Metabolic
disorder}\\
{\em action potential}&{\em Evoked potential}\\
{\em action potential}&{\em Membrane potential}\\
{\em activity curve}&{\em Time curve}\\
\hline
\end{tabular}
}
\caption{\label{tabIns}Examples of term acquisition through insertion
from [Medic]. Terms which do not belong to the reference list are
underlined.}
\end{table}

\begin{table}[t]
\centering
\mbox{
\begin{tabular}[t]{|ll|}
\hline
Candidate term&Reference Term\\
\hline
\hline
{\em [of] accessory cell}&{\em Cell proliferation}\\
$\underline{\textrm{\em [in] acetabular growth}}$&{\em Growth
factor}\\
$\underline{\textrm{\em [of] activated b cell}}$&{\em Cell
differentiation}\\
{\em [of] acute phase protein}&{\em Protein synthesis}\\
{\em [of] adipose tissue}&{\em Tissue extract}\\
$\underline{\textrm{\em [in] adult cell}}$&{\em Cell function}\\
$\underline{\textrm{\em [in] agarose gel}}$&{\em Gel
electrophoresis}\\
$\underline{\textrm{\em [of] airway control}}$&{\em Control method}\\
$\underline{\textrm{\em [of] anaphylatoxin level}}$&{\em Level
measurement}\\
$\underline{\textrm{\em [of] aneuploid tumor cell}}$&{\em Cell
population}\\
$\underline{\textrm{\em [of] animal tolerance}}$&{\em Tolerance
limit}\\
{\em [of] arterial pressure}&{\em Pressure control}\\
\hline
\end{tabular}
}
\caption{\label{tabPerm}Examples of term acquisition through
permutation from [Medic]. Terms which do not belong to the reference
list are underlined.}
\end{table}

\subsection*{Coordination}

Two terms are coordinated only if they share the same semantic scheme.
For example, the variant {\em surgical exploration and closure} (see
the first example of Table~\ref{tabacqui}) indicates that the two
terms {\em surgical exploration} and {\em surgical closure} are
semantically close. They both denote a surgical act. This fact is
interesting because some of the terms with a {\em
surgical\,$\langle$Noun$\rangle$} structure such as {\em surgical
shock} do not belong to the same conceptual class and could not be
coordinated with any of the {\em surgical\,$\langle$Noun$\rangle$}
terms from this class~: *{\em a surgical shock and closure} is
incorrect. Thus, when heads are coordinated (approximately 15\,\% of
the coordinations) the head nouns of the terms must belong  to the
same semantic class (with respect to their entry selected by their
argument). On the other hand, when arguments are coordinated, they
must select the same entry of the head noun. For example, {\em dorsal
spine} and {\em cervical spine} can be coordinated as both being a
part of the {\em (nervous) spine} but neither of them can be
coordinated with a {\em hedgehog} or a {\em fish spine}. Such
coordinations are useful indicators for the disambiguation of a head
word by its arguments~:
\begin{itemize}
\item
For its classification with other related words through head
coordination.
\item
For the definition of its subsenses depending on its arguments through
argument coordination.
\end{itemize}

This kind of fine-grained selectional restriction has to be completed
with more general information on argument structure through long
distance dependencies. Such restrictions can be acquired from
statistical measures on the results of a shallow syntactic analysis
and semantic tags, whether manually assigned \cite{BasEtAl93} or
deduced from a thesaurus \cite{Resnik93}. These studies provide more
general and systematic restrictions than our approach and are applied
to disambiguation or parsing tasks. Our acquisition is restricted to
local selection but takes advantage of the pre-existing knowledge
embodied in lists of reference terms.

\begin{figure*}
\centering
\mbox{\em    \setlength{\unitlength}{0.92pt}
\begin{picture}(487,300)
\thinlines    \put(10,10){\framebox(477,290){}}
              \put(452,96){\vector(-4,-1){238}}
              \put(317,265){\vector(0,1){13}}
              \put(237,100){\vector(0,-1){13}}
              \put(82,134){control}
              \put(82,149){Uraemic}
              \put(255,161){control}
              \put(255,176){Middle aged}
              \put(67,253){Normal control}
              \put(292,253){Disease control}
              \put(82,110){Solvent control}
              \put(103,250){\vector(0,-1){88}}
              \put(103,250){\vector(-3,-1){62}}
              \put(103,250){\vector(1,-3){43}}
              \put(151,258){\vector(1,0){133}}
              \put(284,254){\vector(-1,0){133}}
              \put(166,218){Matched control}
              \put(317,219){Age matched control}
              \put(289,282){Nondemanded control}
              \put(260,136){Weight matched control}
              \put(215,105){Pair fed control}
              \put(17,218){Cirrhotic}
              \put(17,153){patient}
              \put(17,203){control}
              \put(17,168){Cirrhotic}
              \put(40,199){\vector(0,-1){15}}
              \put(316,250){\vector(-4,-1){82}}
              \put(316,250){\vector(3,-1){54}}
              \put(203,213){\vector(-1,-3){42}}
              \put(203,213){\vector(3,-1){82}}
              \put(203,213){\vector(0,-1){176}}
              \put(334,152){\vector(-2,1){35}}
              \put(373,213){\vector(-1,-2){32}}
              \put(203,213){\vector(1,-3){31}}
              \put(155,22){Age matched healthy control}
              \put(213,76){Ad libitum control}
              \put(382,100){Sex matched control}
              \put(390,213){\vector(2,-3){66}}
              \put(431,285){Gender}
              \put(431,270){matched}
              \put(431,255){control}
              \put(431,210){Race}
              \put(431,195){matched}
              \put(431,180){control}
              \put(389,233){\vector(1,1){35}}
              \put(400,215){\vector(2,-1){25}}
              \put(140,75){Young}
              \put(140,60){healthy}
              \put(140,45){control}
              \put(448,114){\vector(-2,3){66}}
\end{picture}}
\caption{\label{figresCoor}Network of coordination links from
[Medic].}
\end{figure*}
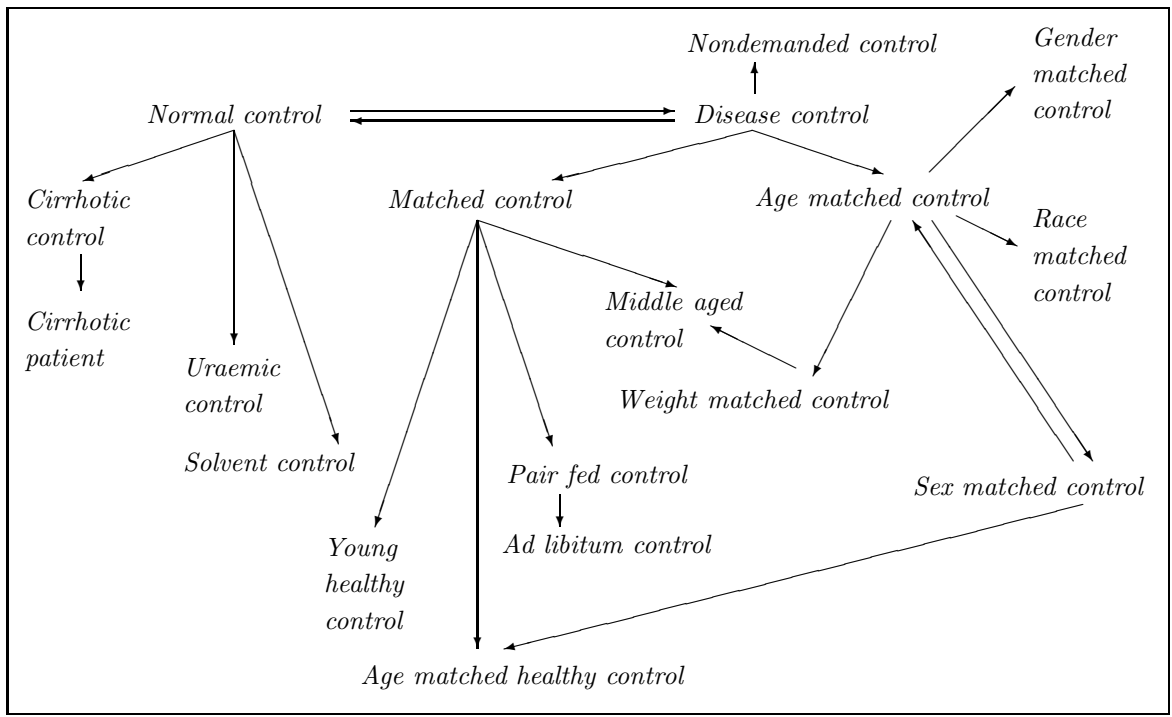

The acquisition from variants, illustrated for one step 
in Table~\ref{tabacqui}, is repeated on
candidate terms as long as new candidates are discovered. Then classes
of compatible sense restrictions are built from terms related through
constructions of coordination according to the following rule~:
\begin{quote}
Two terms $t$ and $t'$ are placed in the same class if and only if 
there exists a chain of coordination variants from $t$ to $t'$~: a set
of $n$ terms $t_1=t$, $t_2,\ldots,$ $t_{n-1}$, $t_n=t'$ such that for
each pair $(t_i, t_{i+1})_{i \in \left\{ 1, 2, \ldots ,n-1 \right\} }$
either $t_i$ is acquired from a coordination variant of $t_{i+1}$ or
$t_{i+1}$ is acquired from a coordination variant of $t_i$.
\end{quote}
Figure~\ref{figresCoor} is a planar representation of the graph
constructed from one of the classes observed in the [Medic] corpus.
Each arrow from a
term $t$ to a term $t'$ indicates that $t'$ has been acquired from a
coordination variant of $t$.

Leaving apart the only head coordination in the figure that holds
between {\em cirrhotic control} and {\em cirrhotic patient}, all the
terms have a {\em $\langle$Modifier$\rangle$\,control}
structure\,\footnotemark\footnotetext{{\em Matched control} is a
partial term with a missing noun argument which is not ruled out by
our acquisition process. With a proper acquisition, this term would
not appear as a candidate and the links issuing from this term would
issue from one of the correct terms {\em
$\langle$Noun$\rangle$\,matched control}.} and can be coordinated
through a head coordination. Conceptually, the terms of
Figure~\ref{figresCoor} are related to a common hypernym whose
linguistic utterance is {\em medical control}.

Moreover, the spatial organization of the graph outlines the central
role played by {\em normal control} and {\em disease control}. These
two terms are the most generic ones. Their root position in this
acyclic graph (except for the two symmetric links) mirrors the
linguistic fact that an argument coordination between two terms tends
to place first the most generic argument and then the most specific
one. Thus, although placed at a similar conceptual level in the
taxonomy, these terms are ordered from the most generic to the most
specific along the coordination links. This two-level observation
reveals that linguistic clues, when precisely observed, are good
indications of the conceptual organization.

\subsection*{Insertion}

\stepcounter{equation}\newcounter{formins}\setcounter{formins}{\theequation}
The meta-rules accounting for insertions insert one or more words
inside a term string. The following meta-rule~(\theformins) denotes an
insertion of one word inside a two-word term~:

\vskip 3mm
{
{\begin{tabular}[t]{ll}
Metarule&$ Ins(X_1 \rightarrow X_2\ X_3)$\\
&$\equiv\ \ X_1 \rightarrow X_2\ X_4\ X_3\ :.$
\end{tabular}}
\hfill(\theformins)
}
\vskip 3mm

\noindent The resulting structure is ambiguous depending on whether
the leftmost word of the term is still an argument of the head noun in
the variation (e.g. {\em [inflammatory [bowel disease]]}\,) or an
argument of the inserted word (e.g. {\em [[sunflower seed] oil]}\,).
The second structure is quite rare and does not correspond to a
genuine variant of the original term because it has a different argument
structure. However, most of these possibly incorrect variants are
correct. It happens every time when the reference term (here {\em
sunflower oil}) corresponds to an elided denomination of the variant
which is in fact the reference term. In this case, the non-ambiguity
of the elided form relies on pragmatic knowledge, because everyone
knows that the {\em seed} is the part of the {\em sunflower} used to
make {\em sunflower oil}.

Whatever the structure of the variant, either $((X_3\ X_2)\ X_1)$\ or
$(X_3\ (X_2\ X_1))$,\ the extraction of the sequence $X_2\ X_1$ as
candidate term (see Tables~\ref{tabacqui} and~\ref{tabIns}) yields a
correct term. When extracted from the latter structure, the candidate
term is more specific than the original one because modifiers in the
noun phrase tend to be ordered from the most generic to the most
specific.

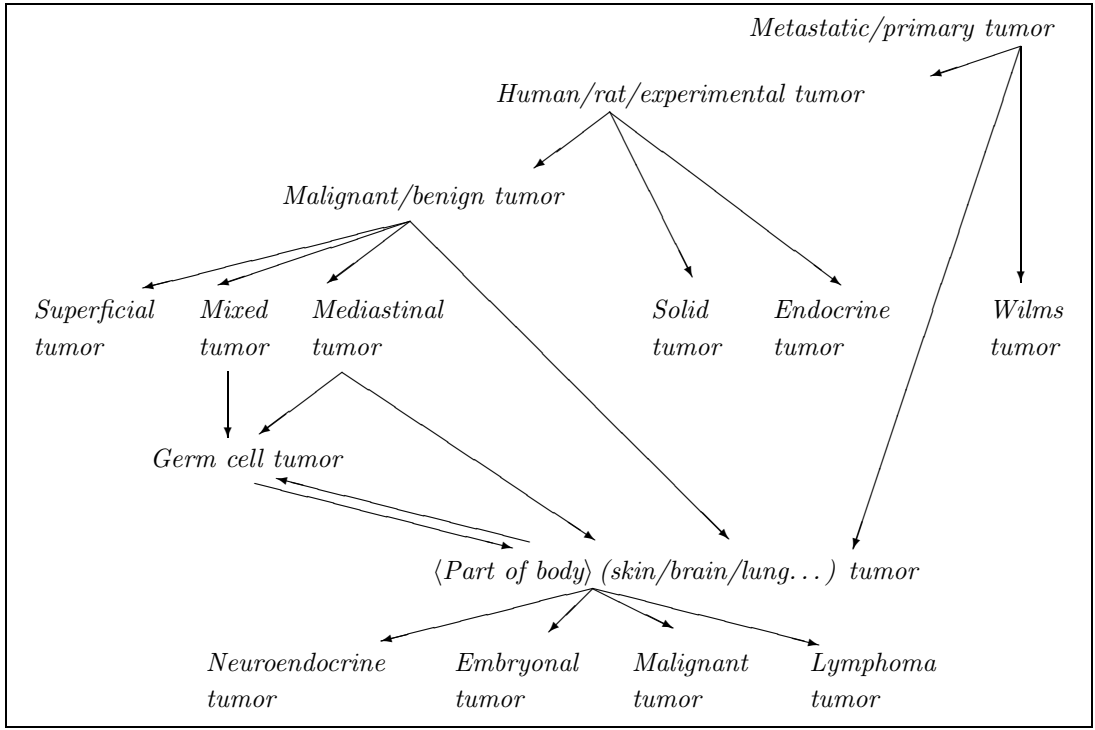
\begin{figure*}
\centering
\mbox{\em    \setlength{\unitlength}{0.92pt}
\begin{picture}(456,307)
\thinlines    \put(10,10){\framebox(446,297){}}
              \put(148,156){\vector(3,-2){104}}
              \put(148,156){\vector(-4,-3){34}}
              \put(251,67){\vector(-4,-1){87}}
              \put(251,67){\vector(-1,-1){18}}
              \put(251,67){\vector(4,-1){93}}
              \put(251,67){\vector(2,-1){33}}
              \put(225,86){\vector(-4,1){104}}
              \put(112,110){\vector(4,-1){106}}
              \put(101,156){\vector(0,-1){27}}
              \put(176,218){\vector(-4,-1){110}}
              \put(176,218){\vector(-3,-1){79}}
              \put(176,218){\vector(-4,-3){34}}
              \put(176,218){\vector(1,-1){131}}
              \put(258,263){\vector(-4,-3){31}}
              \put(427,290){\vector(0,-1){97}}
              \put(427,290){\vector(-3,-1){37}}
              \put(427,290){\vector(-1,-3){69}}
              \put(258,263){\vector(4,-3){95}}
              \put(258,263){\vector(1,-2){34}}
              \put(267,18){tumor}
              \put(267,33){Malignant}
              \put(340,18){tumor}
              \put(340,33){Lymphoma}
              \put(92,18){tumor}
              \put(92,33){Neuroendocrine}
              \put(194,18){tumor}
              \put(194,33){Embryonal}
              \put(414,163){tumor}
              \put(414,178){Wilms}
              \put(275,163){tumor}
              \put(275,178){Solid}
              \put(325,163){tumor}
              \put(325,178){Endocrine}
              \put(69,117){Germ cell tumor}
              \put(21,163){tumor}
              \put(21,178){Superficial}
              \put(89,178){Mixed}
              \put(89,163){tumor}
              \put(135,163){tumor}
              \put(135,178){Mediastinal}
              \put(185,71){$\langle$Part of
body$\rangle$\,(skin/brain/lung\ldots) tumor}
              \put(123,225){Malignant/benign tumor}
              \put(211,267){Human/rat/experimental tumor}
              \put(315,294){Metastatic/primary tumor}
\end{picture}}
\caption{\label{figresIns}Network of insertion links from [Medic].}
\end{figure*}

As stated for coordination, iteration of acquisition on candidates
terms yields conceptual classes. However, the construction of the
graph linking terms acquired through insertion is not as
straightforward as it is for coordination. The reason is that one
must first conflate conceptually close terms that are likely to be
coordinated before constructing the hierarchy resulting from insertion
variants. Figure~\ref{figresIns} has been constructed by grouping
together {\em malignant tumor/benign tumor}, {\em metastatic
tumor/primary tumor} and {\em human tumor/experimental tumor} which
have been observed in coordinated constructions. A further grouping of
{\em rat tumor} with {\em human tumor} was necessary but was not
indicated by a coordination in our corpus. Similarly, a general
category of {\em $\langle$Part of body$\rangle$\,tumors} has been
created although only some coordinations were observed among the
possible ones~: {\em mammary/skin, mammary/pancreas,
cutaneous/corneal, liver/lung, bone/soft tissue}\ldots

Due to the parallel between insertion constructions and
generic/specific links, there is a good similarity between the
observed graph and the taxonomy of this part of the terminology. An
exception to this rule is the link from {\em $\langle$Part of
body$\rangle$\,tumor} to {\em malignant tumor} coming from the variant
{\em ovarian malignant tumor}. It is indeed an exceptional link~:
there are fifteen different links from {\em malignant tumor} to more
specific terms but only one link from a more specific term ({\em
ovarian tumor}) to {\em malignant tumor}.

\section{Incrementality and Robustness}

As introduced for the construction of conceptual classes, the
acquisition method is repeated incrementally. Candidates are acquired
from candidates of the preceding step until no new term is
discovered~: 
\begin{quote}
A term is a candidate if and only if there exists a chain of couples
$(t_i,t_{i+1})_{i \in \left\{ 1, 2, \ldots ,n-1 \right\} }$\ where
$t_{i+1}$\ is acquired from a variant of $t_i$\ and where $t_1$\ is a
reference term. That is to say that the set of candidates is the
closure of the set of reference terms through the relation of
acquisition.
\end{quote}
Due to the finite corpus, due to the finite length of terms and
due to the non circularity of the definition, the
incremental acquisition reaches a fixed point after a finite number of
iterations. It takes fifteen cycles to complete an acquisition of
5,080 terms when starting from the 71,623 terms of the [Pascal]
list\,\footnotemark\footnotetext{Among these 71,623 terms, only 12,717
are found in the [Medic] corpus under their basic form or one of its
correct variants.}.

Table~\ref{tabchAcq} shows five sequences of acquisition obtained from
term variants in [Medic] starting from a reference term in [Pascal].
For example, the first sequence indicates the acquisition of {\em
tumour tissue} from {\em tissue extract} through a permutation variant
({\em extract of tumour tissue}) followed by the acquisition of {\em
normal tissue} from a coordination ({\em tumour or normal tissue}),
and so on. This sequence mixes the three kinds of variations while the
last three are restricted to insertions and/or coordinations. When not
using permutation, the acquisition process yields smaller sets of
terms~: it produces 2,998 terms in fourteen steps through
coordinations and insertions, 2,193 terms in seven steps through
insertions and 357 terms in six steps through coordinations. The sets
obtained without the use of permutation are ``better'' candidates
because they are produced by transformations which yield compounds.
Permutations, which transform compounds into syntactic noun phrases,
tend to produce candidates of a lower quality.

\begin{table}[t]
\centering
\mbox{
\em
\begin{tabular}[t]{|l|ll|}
                            \hline
{\em Var}&\multicolumn{2}{c|}{{\em Acquired terms}}\\
                            \hline
                            \hline
{\em P}&Tissue extract&tumour tissue\\
{\em C-I}&normal tissue&rat tissue\\
{\em P-I}&sprague dawley rat&female rat\\
{\em I-I}&f344 rat&strain rat\\
{\em P-I}&milan strain&normotens. strain\\
{\em I}&control strain&\\
                            \hline
                            \hline
{\em I}&Blood cell&leukemic cell\\
{\em C-C}&normal cell&cf cell\\
{\em I-I}&pancreatic cell&beta cell\\
{\em C-I}&alpha cell&activated NK cell\\
                            \hline
                            \hline
{\em I}&Cell line&tumor line\\
{\em I-I}&derived cell line&t cell line\\
{\em I-I}&leukemia cell line&u937 cell line\\
{\em I}&histiocytic cell line&\\
                            \hline
                            \hline
{\em C}&Experimental study&clinical study\\
{\em C-C}&echocardiogr. study&doppler study\\
{\em C}&angiography study&\\
                            \hline
                            \hline
{\em C}&Pigment. disorder&nail disorder\\
{\em C-C}&nail change&palmar change\\
                            \hline
\end{tabular}
}
\caption{\label{tabchAcq}Examples of sequences of acquisition.}
\end{table}

\begin{figure}[t]
\centering
\mbox{
\begin{picture}(231,126)
\includegraphics{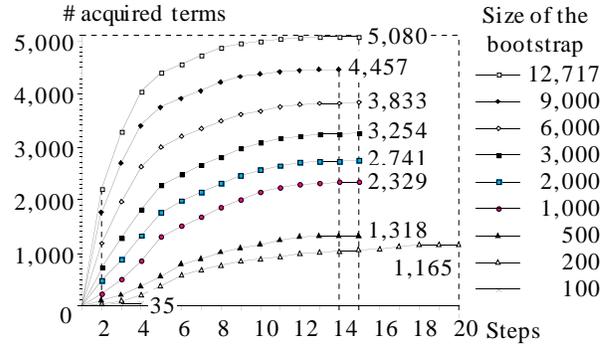}
\end{picture}
}
\caption{\label{figroleBootstr}Acquisition volumes for different sizes
of bootstrap on [Medic] corpus.}
\end{figure}

As our method is based on the observation of rare occurrences, the
number of acquired terms depends on the set of reference terms. As
indicated in \cite{Eng94}, such a correlation does not exist in her
statistical approach to term acquisition because she observes larger
sets of (co-)occurrences. Figure~\ref{figroleBootstr} exemplifies
acquisition curves for different values of the volume of reference
terms. It shows that the size of the acquisition gradually degrades
when the size of the bootstrap decreases~: 5,080 terms are acquired
when starting from the total list of 12,717 terms, 3,833 terms are
still acquired from a bootstrap of 6,000 terms and 2,329 terms from a
bootstrap of 1,000 terms. Thus, with only a twelfth of the initial
bootstrap, almost half the terms are still acquired. Although a
serious degradation of the results is observed under this lower limit,
these values suggest that acquisition depends more on the size of the
corpus than on the initial terminology. As a partial initial list of
terms is easily compensated by a larger corpus, the completeness of
the reference list is not a crucial issue for the quality of the
acquisition in our framework.

\section{Conclusion and Future Work}

This study has proposed a novel approach to terminological acquisition
that differs from the two main trends in this domain~:
morpho-syntactic filtering or statistical extraction. The main feature
of our approach is accounting for existing lists of terms by observing
their variants and yielding conceptual links as well as candidate
terms. As long as they are accessible through morpho-syntactic
dependencies in a corpus, these links can be used to  automatically
construct parts of the taxonomy representing the knowledge in this
domain. Among the applications of this method are lexical acquisition,
thesaurus discovery and technological survey. More generally,
terminological enrichment is necessary for NLP activities dealing with
technical sublanguages because their efficiency and their quality
depend on the completeness of their lexicons of terms and compounds.

\newcommand{\noopsort}[1]{} \newcommand{\printfirst}[2]{#1}
  \newcommand{\singleletter}[1]{#1} \newcommand{\switchargs}[2]{#2#1}

\bibliographystyle{aaai}

\end{document}